\documentclass[sigplan,10pt,screen]{acmart}
\usepackage{microtype}
\usepackage{graphicx}
\usepackage{booktabs} 
\usepackage{amsfonts}
\usepackage{amsmath}
\usepackage{listings}
\usepackage{xcolor}
\usepackage{float}
\usepackage{tabularx} 
\usepackage{caption}
\usepackage{subcaption}
\usepackage{adjustbox}
\usepackage{enumitem}
\usepackage{tikz}
\usepackage{pict2e,circledsteps,pgfkeys}
\usepackage{multirow}
\usepackage{geometry}
\usepackage{hyperref}
\usepackage{array}
\usepackage{colortbl}
\usepackage{xurl}
\usepackage{color,soul}
\usepackage{xspace}
\usepackage{comment}
\usepackage{tcolorbox} 
\tcbuselibrary{listings}      

\definecolor{pblue}{RGB}{78,121,167}
\definecolor{pred}{RGB}{225,87,89}
\definecolor{pgreen}{RGB}{89,161,79}
\definecolor{darkgreen}{rgb}{0,0.5,0}

\newcommand*\wcircled[1]{\tikz[baseline=(char.base)]{
    \node[shape=circle,draw=black,inner sep=0.5pt] (char) {\textcolor{black}{\textit{#1}}};}}

\lstdefinestyle{xmlstyle}{
  basicstyle=\ttfamily\scriptsize,
  morestring=[s]{"}{"},
  morecomment=[s]{?}{?},
  morecomment=[s]{!--}{--},
  commentstyle=\color{darkgreen},
  moredelim=[s][\color{black}]{>}{<},
  moredelim=[s][\color{pred}]{\ }{=},
  stringstyle=\color{pgreen},
  identifierstyle=\color{pblue},
  tabsize=2,
  breaklines=true,
  frame=single,
  xleftmargin=3.4pt,
  xrightmargin=3.4pt
}

\newtcblisting{examplecode}{
  listing only,
  colback=gray!5,
  colframe=white,
  boxrule=0pt,
  arc=0pt,
  top=-3pt, bottom=-3pt, left=16pt, right=1pt,
  width=\linewidth,
  listing options={
    basicstyle=\ttfamily\small,
    keywordstyle=\color{blue}\bfseries,
    keywords={forward,synchronize,create_queue,mask,copy,embed_txt, tokenize, detokenize, alloc_emb, dealloc_emb, alloc_kvpage, dealloc_kvpage, get_next_dist},
    breaklines=true,
    numbers=left,
    numberstyle=\small,
    numbersep=8pt,
    showstringspaces=false,
    tabsize=2,
    lineskip=-3pt,
    breakindent=10pt,        
    breakautoindent=true   
  }
}

\newcommand{\code}[1]{\texttt{\small#1}\xspace}

\newcommand{\paragraphb}[1]{\vspace{0.05in}\noindent{\bf #1.}}

\newcommand{\sys}{{{\sc Pie}}\xspace}
\newcommand{\Program}{Inferlet\xspace}
\newcommand{\Programs}{Inferlets\xspace}
\newcommand{\program}{inferlet\xspace}
\newcommand{\programs}{inferlets\xspace}

\newcommand{\embed}{{\sc{Embed}}\xspace}
\newcommand{\kv}{{\sc{KvPage}}\xspace}
\newcommand{\embeds}{{\sc{Embeds}}\xspace}
\newcommand{\kvs}{{\sc{KvPages}}\xspace}
\newcommand{\cmdqueue}{{\sc{Queue}}\xspace}
\newcommand{\cmdqueues}{{\sc{Queues}}\xspace}

\def\ie{{i.e.}}
\def\eg{{\em e.g.}\xspace}

\definecolor{lightgray}{gray}{0.9}
\definecolor{lightblue}{rgb}{0.9,0.9,1}
\definecolor{blue_bg}{rgb}{0.7,0.85,1}
\definecolor{lightyellow}{rgb}{1,1,0.8}
\definecolor{lightpurple}{rgb}{1,0.85,1}
\definecolor{red}{rgb}{1,0,0}
\definecolor{darkgreen}{rgb}{0.4,0.7,0.3}
\definecolor{darkblue}{rgb}{0.2,0.7,0.9}

\AtBeginDocument{%
  }

\copyrightyear{2025}
\acmYear{2025}
\setcopyright{cc}
\setcctype{by}
\acmConference[SOSP '25]{ACM SIGOPS 31st Symposium on Operating Systems Principles}{October 13--16, 2025}{Seoul, Republic of Korea}
\acmBooktitle{ACM SIGOPS 31st Symposium on Operating Systems Principles (SOSP '25), October 13--16, 2025, Seoul, Republic of Korea}\acmDOI{10.1145/3731569.3764814}
\acmISBN{979-8-4007-1870-0/2025/10}

\acmBadgeR[https://www.acm.org/publications/policies/artifact-review-and-badging-current]{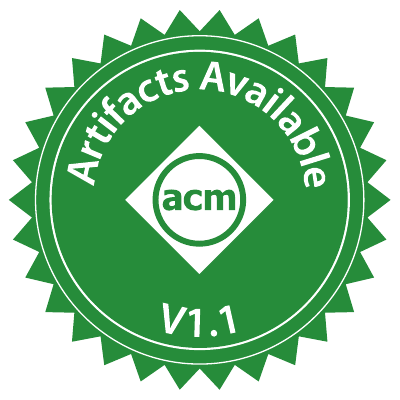}
\acmBadgeR[https://www.acm.org/publications/policies/artifact-review-and-badging-current]{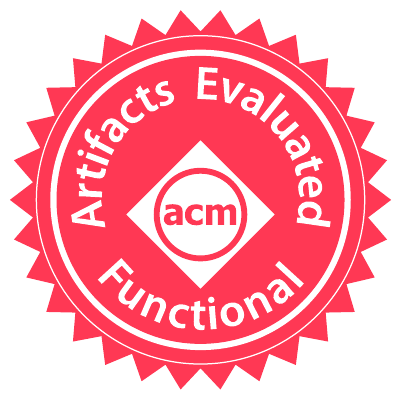}
\acmBadgeR[https://www.acm.org/publications/policies/artifact-review-and-badging-current]{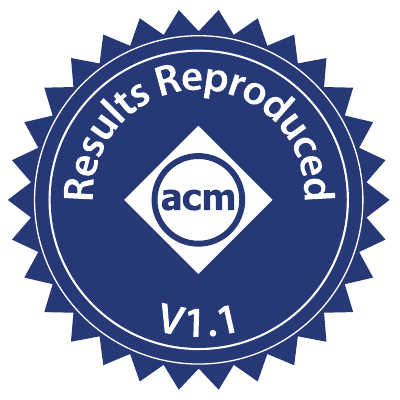}

\begin{document}
\sloppy
\title[Pie: A Programmable Serving System for Emerging LLM Applications]{Pie: A Programmable Serving System\\ for Emerging LLM Applications}

\settopmatter{authorsperrow=4}

\author{In Gim} 
\affiliation{%
  \institution{Yale University}
  \city{ }
  \country{ }
}

\author{Zhiyao Ma}
\affiliation{%
  \institution{Yale University}
  \city{ }
  \country{ }
}

\author{Seung-seob Lee}
\affiliation{%
  \institution{Yale University}
  \city{ }
  \country{ }
}

\author{Lin Zhong}
\affiliation{%
  \institution{Yale University}
  \city{ }
  \country{ }
}

\renewcommand{\shortauthors}{Gim et al.}

\begin{abstract}
Emerging large language model (LLM) applications involve diverse reasoning strategies and agentic workflows, straining the capabilities of existing serving systems built on a monolithic token generation loop. This paper introduces \sys, a programmable LLM serving system designed for flexibility and efficiency. \sys decomposes the traditional generation loop into fine-grained service handlers exposed via an API and delegates control of the generation process to user-provided programs, called \emph{\programs}. This enables applications to implement new KV cache strategies, bespoke generation logic, and seamlessly integrate computation and I/O---entirely within the application, without requiring modifications to the serving system. \sys executes \programs using WebAssembly, benefiting from its lightweight sandboxing. Our evaluation shows \sys matches state-of-the-art performance on standard tasks (3-12\% latency overhead) while significantly improving latency and throughput (1.3$\times$-3.4$\times$ higher) on agentic workflows by enabling application-specific optimizations.
\end{abstract}





\maketitle

\section{Introduction}
\label{sec:introduction}

Large language model (LLM) usages are evolving far beyond simple text completion. Today's LLM serving systems require sophisticated strategies to execute LLM application logic ranging from token-level~\cite{beurer-kellner2024guiding,holtzman2020the,park2024grammar} to workflow-level~\cite{ye2024chunkattention,gim2024prompt,zheng2024sglang}. These systems must also handle real-time user events, such as audio~\cite{bai2025qwen2} and vision~\cite{team2024gemma}, while seamlessly integrating with external tools~\cite{abhyankar2024infercept,yu2025stateful,gao2025fast} and code execution environments~\cite{wang2024executable,trinh2024solving,ni2023lever} to support increasingly common agentic workflows~\cite{wang2024survey,compound-ai-blog}.

These emerging LLM applications reveal fundamental limitations in the efficiency and flexibility of today's serving infrastructure (\S\ref{sec:background}), prompting us to identify three requirements for next-generation LLM serving systems:

\begin{enumerate}[leftmargin=*,topsep=0.3em,itemsep=0.15em, label=R\arabic*.]
    \item \textbf{Application-specific KV cache control.} LLM workflows such as tree-of-thought reasoning~\cite{yao2023tree}, map-reduce summaries~\cite{besta2024graph}, multi-step generations~\cite{khattab2022demonstrate}  require explicit, fine-grained control over KV cache management. This includes customizing allocation~\cite{kwon2023efficient}, eviction policies~\cite{yu2025stateful}, and reuse strategies~\cite{zheng2024sglang} according to application-specific logic, rather than relying on the implicit, system-wide heuristics built into serving systems.
    \item \textbf{Customizable generation processes.} Emerging decoding methods (assisted decoding~\cite{dong2024xgrammar}, MCTS-based generation~\cite{gupta2024language}), stateful sampling strategies~\cite{holtzman2020the}, and safety grounding~\cite{kumar2023certifying} require the ability to precisely control and potentially modify the token prediction and sampling loop on a per-request or even per-step basis.
    \item \textbf{Integrated computation and I/O.} Agentic workflows and interactions with external systems necessitate tightly coupling token generation with arbitrary computations (\eg, running symbolic checks~\cite{trinh2024solving}) and I/O operations (\eg, making API calls~\cite{schick2023toolformer, suris2023viper}) within the generation flow, without incurring prohibitive latency penalties or complex external orchestration.
\end{enumerate}

\begin{figure}[t]
  \centering
  \includegraphics[width=\columnwidth]{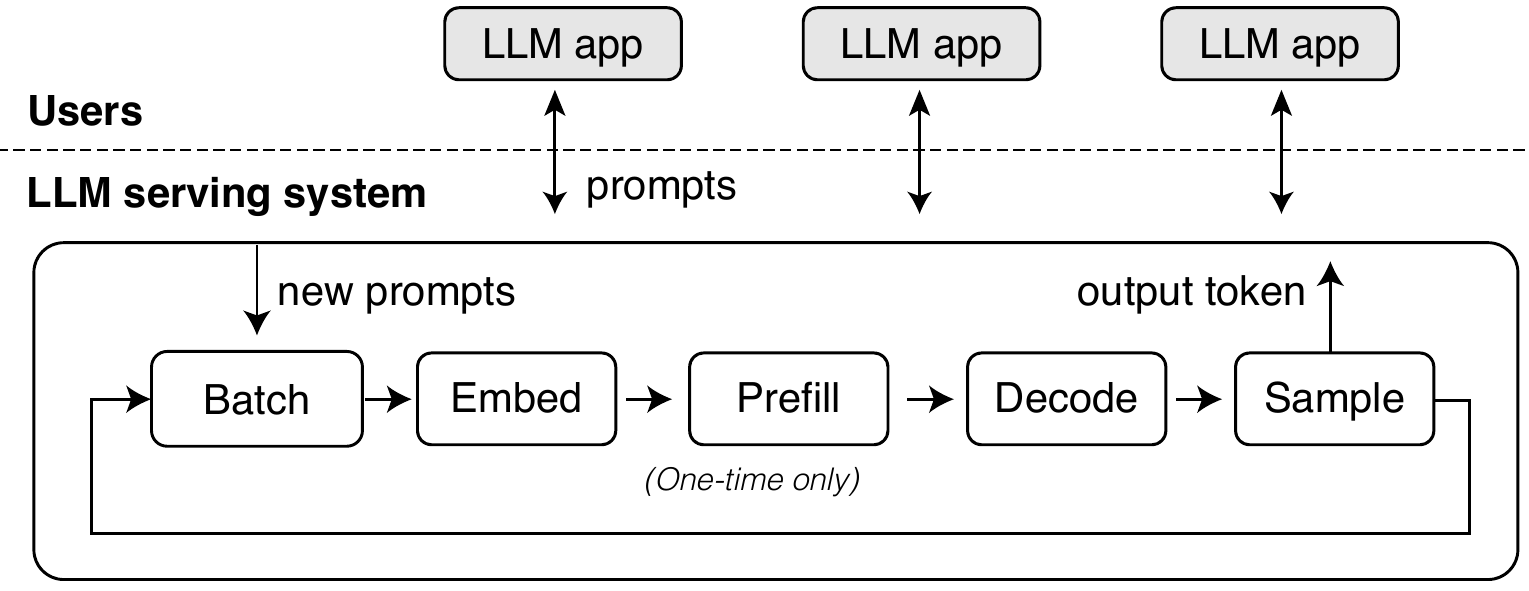}
  \caption{\textbf{Current LLM serving systems} conceptually follow a monolithic prefill-decode loop, batching prompts and applying global policies for KV cache. This design lacks flexibility to support application-specific logic.}
  \label{fig:arch-existing-systems}
  \vspace{-3pt}
\end{figure}

Existing systems, such as vLLM~\cite{kwon2023efficient} and TGI~\cite{huggingface_text_generation_inference}, struggle to meet these requirements due to their reliance on an inflexible, monolithic token generation loop (\autoref{fig:arch-existing-systems}). This architecture enforces global policies (\eg, for KV cache management, hindering R1), employs a closed generation process that resists customization (hindering R2), and isolates inference from external operations, such as tool use, forcing the client to coordinate such operations or otherwise requiring custom modifications to the serving system (hindering R3). While recent systems like SGLang~\cite{zheng2024sglang} and Parrot~\cite{lin2024parrot} introduce some programmatic abstractions, they are fundamentally constrained by the same monolithic design (\S\ref{sec:related_work}).

\begin{figure}[t]
\includegraphics[width=\columnwidth]{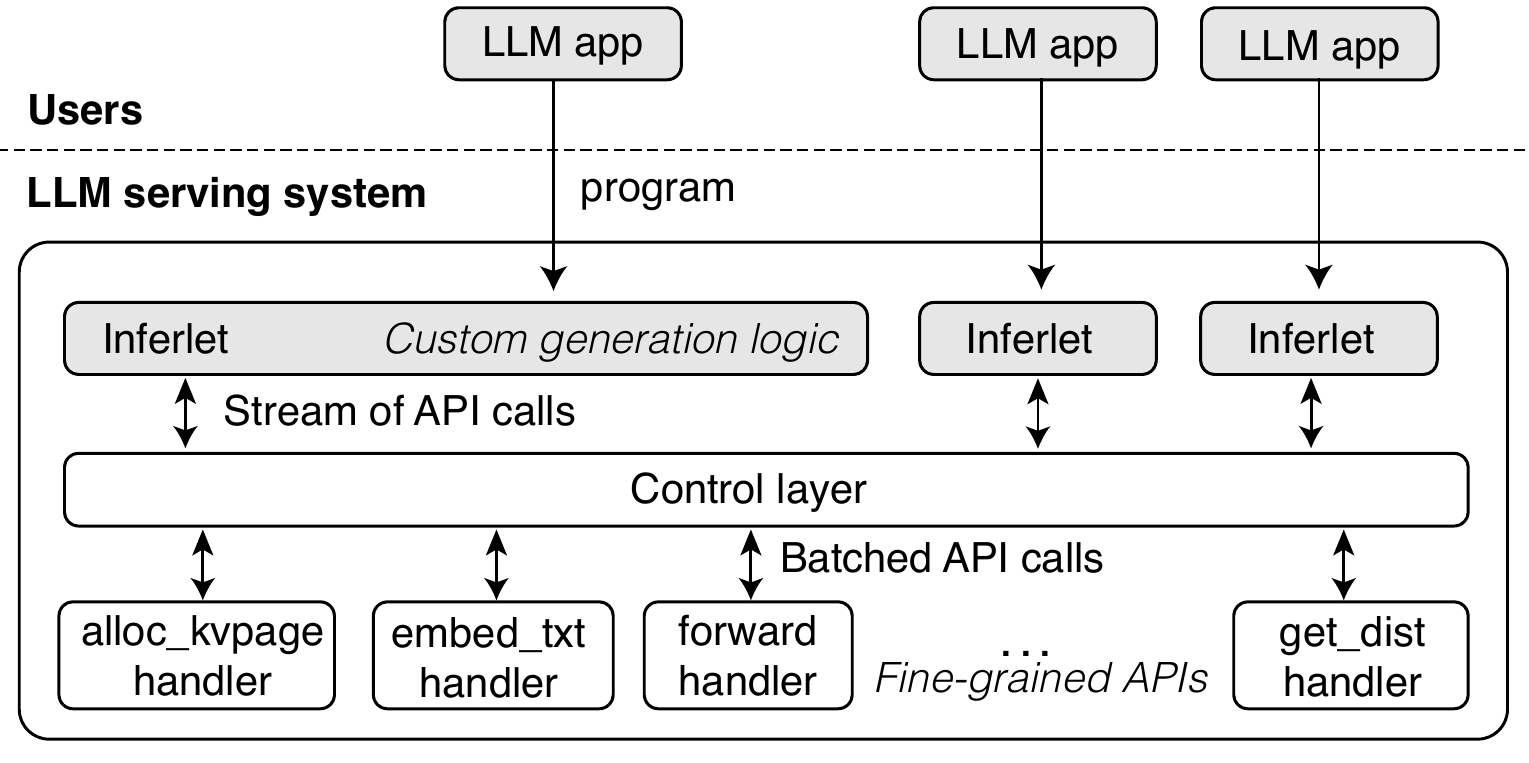}
  \caption{\textbf{Our proposed system, \sys}, dismantles the sequential generation process into independent handlers, and delegates control to user-provided programs called \emph{\programs}.}
  \label{fig:arch-proposed-system}
\end{figure}

This paper introduces \emph{\sys}, a programmable LLM serving system designed to meet these requirements. The core idea of \sys is to \emph{delegate control over the end-to-end generation process to user-provided programs}, called \emph{\programs}. \sys decomposes the process into fine-grained handlers for stages like embedding, forward pass, and sampling (\autoref{fig:arch-proposed-system}), instead of a global generation pipeline. \Programs orchestrate these handlers via API calls (\S\ref{sec:api}), enabling them to explicitly manage resources like the KV cache, define custom generation logic, and directly integrate arbitrary computation and I/O.

Instead of treating prompts as the basic unit of service, \sys elevates programs to this role, allowing hundreds of concurrent \programs to adopt distinct optimization strategies. For example, one may exploit custom KV caching, another speculative decoding~\cite{leviathan2023fast}, and yet a third may exploit a complex agentic loop, all on the same underlying serving infrastructure as illustrated by \autoref{fig:arch-proposed-system}. Viewed through this lens, all existing serving systems effectively operate with a single, fixed \program, \ie, an autoregressive loop, which explains why they are inflexible for modern AI applications.

We implement \sys using a layered architecture (\S\ref{sec:arch}) and build its \program runtime around WebAssembly (Wasm). Developers can program \programs using any programming language that compiles to Wasm (\eg, C++, Rust, Python) with the API bindings provided by \sys.

Our evaluation (\S\ref{sec:evaluation}) demonstrates the effectiveness of \sys. We implement a diverse range of LLM techniques as \programs, including attention variants~\cite{xiao2024efficient,beltagy2020longformer}, constrained and speculative decoding~\cite{santilli2023accelerating,beurer-kellner2024guiding}, deliberate prompting strategies~\cite{yao2023tree,besta2024graph}, and agentic workflows. We show that \sys matches state-of-the-art performance for traditional tasks (\eg, 3-12\% latency overhead for text completion) while significantly outperforming existing systems on advanced tasks like Graph-of-Thought and agentic workflows (1.1$\times$-2.4$\times$ lower latency, 1.3$\times$-3.4$\times$ higher throughput) by enabling application-specific optimizations.

\vspace{0.7em}

In summary, we report the following contributions.
\begin{itemize}[leftmargin=*,topsep=0.3em,itemsep=0.15em, label=$-$]
    \item Identifying three key limitations in current LLM application serving: implicit resource control, inflexible generation processes, and poor workflow integration.

    \item The \sys architecture, which decomposes the monolithic generation pipeline into fine-grained service handlers.

    \item A Turing-complete programming model providing applications with full and end-to-end control over the generative workflow, KV cache, and I/O.

    \item Implementation and an evaluation demonstrating \sys matches state-of-the-art performance on standard tasks while significantly improving latency and throughput on emerging applications.
\end{itemize}
\noindent \sys is open-sourced at \url{https://github.com/pie-project/pie}.

\section{Background and Motivation}
\label{sec:background}

We first outline the architecture underpinning existing LLM serving systems. We then detail how this architecture falls short in supporting increasingly complex LLM applications.

\subsection{How LLM serving systems work}

Existing LLM serving systems~\cite{yu2022orca,sheng2023flexgen,fu2024serverlessllm,kwon2023efficient,huggingface_text_generation_inference,agrawal2024taming} are designed for high-throughput text completion, treating each user request as a single input prompt to be processed by a prefill-and-decode model (\autoref{fig:arch-existing-systems}). This generation process is composed of discrete steps. It begins with a prefill step, where the system processes the entire prompt in parallel to populate an initial Key-Value (KV) cache to avoid redundant computation. The system then enters an iterative decode phase, where each step uses the KV cache and the last token to predict and sample the next token, then updates the cache. To maximize hardware utilization, a central scheduler groups multiple requests (prompts) into a batch, which advances all of them through each generation step in lockstep.

\subsection{Limitations of existing serving systems}

The current LLM serving architecture centers around three key aspects whose inherent inflexibility creates challenges for emerging applications: (1) the management of the KV cache, (2) the fixed structure of the token prediction and sampling loop, and (3) the integration of the generation process with external computations or workflows.

\paragraphb{Implicit KV cache management}
Current systems typically manage the KV cache using implicit, system-wide policies, such as LRU eviction~\cite{zheng2024sglang,kwon2023efficient} or prompt caching~\cite{gim2024prompt,liu2024cachegen}, where common leading token sequences across requests share cached entries. While efficient for simple prompt variations, this approach fails to provide the application-specific control required by advanced strategies. Techniques like Recursion-of-Thought~\cite{lee2023recursion} and Graph-of-Thought~\cite{besta2024graph} demand explicit, runtime decisions about which specific KV cache blocks to retain, discard, reuse, or duplicate based on the application's dynamic state or reasoning structure—control that global heuristics cannot provide.
Furthermore, fine-grained manipulation of the KV cache structure is needed for techniques like attention sink~\cite{xiao2024efficient} and beam search~\cite{freitag2017beam}. Implementing such features within the monolithic architecture often requires invasive modifications deep within the system's memory manager and scheduler. The engineering effort can hinder adoption; for example, support for beam search was considered for removal from vLLM (v0.6.3) due to its complexity~\cite{vllm6226}.

\paragraphb{Inflexible decoding process}
The tightly coupled ``predict-then-sample'' operation within the monolithic loop offers limited flexibility for customizing the core generation algorithm. Emerging decoding methods like parallel decoding~\cite{santilli2023accelerating,gloeckle2024better} or speculative decoding~\cite{leviathan2023fast,saxena2023prompt} deviate from the standard decoding process, often predicting multiple tokens per step or using verification steps. Integrating these is challenging because their variable output granularity interferes with the regular process, which assumes a single token is generated per request in each batching step. Consequently, such techniques are often implemented as system-wide toggles~\cite{zheng2024sglang,huggingface_text_generation_inference,kwon2023efficient}, rather than being flexibly selectable or configured on a per-request basis. This lack of per-request customization is inefficient, as different applications or even different phases within one application might benefit from distinct decoding strategies. 
Similarly, stateful generation strategies (MCTS~\cite{gupta2024language}, grammar-constrained decoding~\cite{beurer-kellner2024guiding,beurer2023prompting}) require managing state across steps, which is cumbersome in the standard stateless loop. Lastly, dynamic control over the output distribution (for watermarking~\cite{kirchenbauer2023watermark}, safety filters~\cite{kumar2023certifying}) requires injecting logic around the sampling step, for which current systems lack clean, application-customizable hooks.

\paragraphb{Poor workflow integration}
The current architecture inherently assumes a ``closed-loop'' generation process, where the system focuses solely on producing tokens based on the initial prompt and the model's internal state. This is ill-suited for agentic workflows~\cite{khattab2022demonstrate,yao2023react} or interactive applications that must integrate token generation with arbitrary computations, such as code execution~\cite{ni2023lever,suris2023viper}, API calls~\cite{patil2024gorilla}, symbolic solvers~\cite{trinh2024solving}.
Integrating such external computations requires an inefficient workaround: the generation process must return control to the client, which performs the external action and then submits a new request with the updated context to resume generation. This round-trip incurs network latency~\cite{abhyankar2024infercept}. More critically, because serving systems are stateless across requests, the state from the initial generation, \ie, the KV cache, is discarded. The system treats the continuation as a new prompt, forcing a costly re-prefill of the interaction history~\cite{gao2025fast,yu2025stateful}. While techniques like prefix caching can mitigate this, they are system-level heuristics rather than a native solution for stateful interaction.
Additionally, incorporating application-specific pre- or post-processing steps (\eg, custom tokenization adjustments) into the request lifecycle is also challenging without introducing overly complex APIs.  

\vspace{0.5em}

Collectively, these limitations reveal a mismatch: The static, monolithic design optimized for batched text completion cannot efficiently or flexibly accommodate the dynamic, heterogeneous, and interactive nature of modern LLM applications.
The root cause lies in the tight coupling of application control logic with the core execution engine within a fixed architectural pattern. Overcoming these challenges requires a paradigm shift towards decoupling the application's control logic from the underlying model execution infrastructure, as embodied by \sys.

\section{Overview of Pie}
\label{sec:overview}

\sys achieves programmability through two fundamental architectural shifts, derived from the insight that application-specific control logic can be efficiently executed \emph{outside} the core inference engine if provided with the right interfaces.
First, \sys \textit{dismantles the monolithic decoding loop into fine-grained procedures}. Instead of a fixed prefill-decode loop, it exposes a set of independent, fine-grained services (called \emph{handlers}) responsible for specific token prediction stages like input text embedding, KV cache operations (\eg, allocation, update), and model forward pass.
Second, \sys \textit{delegates end-to-end control to user-provided programs}, called \emph{\programs}. These \program programs define bespoke generation workflows by issuing API calls to the handlers. \Programs can:

\begin{enumerate}[leftmargin=*,topsep=0.3em,itemsep=0.02em]
  \item \emph{Manage resources explicitly}, particularly allocating, manipulating, and reusing KV cache pages based on application logic. This provides the fine-grained control needed for advanced caching strategies (R1 in \S\ref{sec:introduction}).
  \item \emph{Orchestrate handlers} precisely, defining custom sequences of operations beyond standard decode loop. This addresses the need for customizable generation processes (R2).
  \item \emph{Integrate arbitrary computation and I/O} seamlessly within their control flow (\eg, calling external APIs, running checks). This tackles the poor workflow integration of existing systems (R3).
\end{enumerate}

To realize this programmable model, \sys employs a layered architecture designed for flexibility and efficiency. At a high level, user \programs execute within a dedicated \textit{application layer}, interacting with the system via the API. The \textit{control layer} orchestrates these interactions, managing resources, virtualizing access, and intelligently batching API calls destined for the hardware. Finally, the \textit{inference layer} executes these batched operations, translating them into low-level computations (\eg, GPU kernel invocations) using specialized handlers for different model operations. 

The following sections elaborate on the specifics of this design. Section \ref{sec:api} details the \textit{\sys programming model}, outlining the API primitives that \programs use to orchestrate generation and manage resources. Section \ref{sec:arch} elaborates on the \textit{three-layer system architecture}, explaining how it supports the execution of \programs, manages control flow, and interfaces with the underlying inference hardware.

\section{Programming Model and API}
\label{sec:api}

\begin{table*}[!htbp]
    \small
    \centering
    \caption{\textbf{Overview of APIs for programming \programs (non-exhaustive).}
        \sys provides a total of 42 APIs, with 18 dedicated to LLM execution and the remainder supporting core runtime operations and agentic workflow. To ensure model-agnosticism and extensibility, \sys organizes LLM-related APIs into traits (\S\ref{sec:traits}). API calls that involve a command queue (\code{q}) are processed by the inference layer, while those that do not are directly handled by the control layer.} 
    \vspace{-0.8em}
    \begin{tabular}{@{}lll@{}}
    \toprule
    \textbf{Trait (Supertraits)} & \textbf{Function} & \textbf{Behavior} \\
    \midrule
    
    & \texttt{\textbf{get\_arg}() -> list[str]} & Gets command-line arguments passed during launch. \\
    & \texttt{\textbf{send}(msg)} & Sends a message to the client that launched the \program. \\
    & \texttt{\textbf{receive}() -> future[str]} & Receives messages from the client. \\
        & \texttt{\textbf{http\_get}(url) -> future[str]} & Performs an HTTP GET request to the specified URL. \\
    & \texttt{\textbf{available\_models}() -> list[Model]} & Gets the list of models. \\
    & \texttt{\textbf{available\_traits}(model) -> list[str]} & Gets the list of given model's traits.\\
    & \texttt{\textbf{create\_queue}(model) -> Queue} & Creates a new command queue. \\
    & \texttt{\textbf{synchronize}(q) -> future} & Blocks until the queue finishes execution. \\
    & \texttt{\textbf{set\_queue\_priority}(q, pri)} & Hints the controller which queue to process first. \\

    \midrule
    \multirow{1}{*}{\emph{Allocate}}
        & \texttt{\textbf{export\_kvpage}(kv, name)} & Exports paged KV cache for use in other programs. \\
    & \texttt{\textbf{import\_kvpage}(name) -> list[KvPage]} & Imports the paged KV cache. \\
    & \texttt{\textbf{alloc\_kvpage}(q, size) -> list[KvPage]} & Allocates memory for paged KV cache. \\
    & \texttt{\textbf{dealloc\_kvpage}(q, kv)} & Deallocates memory for paged KV cache. \\
    & \texttt{\textbf{alloc\_emb}(q, size) -> list[Embed]} & Allocates buffer space for embeddings. \\
    & \texttt{\textbf{dealloc\_emb}(q, emb)} & Deallocates buffer space for embeddings. \\
    & \texttt{\textbf{copy\_kvpage}(q, src, dst)} & Copy KV cache contents at token-level. \\

    \midrule
    \multirow{1}{*}{\emph{Forward} (\emph{Allocate})}
    & \texttt{\textbf{forward}(q, ikv, iemb, okv, oemb, mask)} & Transform to KV cache and/or output embeddings. \\
    & \texttt{\textbf{mask\_kvpage}(q, tgt, mask)} & Masks the KV cache in a token-level manner. \\
    \midrule
    \multirow{1}{*}{\emph{InputText} (\emph{Allocate, Forward})}
    & \texttt{\textbf{embed\_txt}(q, tok, pos, embs)} & Embeds text input into embeddings \\

    \midrule
    \multirow{1}{*}{\emph{InputImage} (\emph{Allocate, Forward})}
        & \texttt{\textbf{num\_embs\_needed}(m, size) -> int} & Calculates the number of embeddings needed. \\

    & \texttt{\textbf{embed\_img}(q, blob, embs)} & Embeds image input into embeddings. \\
    
    \midrule
    \multirow{1}{*}{\emph{Tokenize} (\emph{InputText})}
    & \texttt{\textbf{tokenize}(q, text) -> list[int]} & Converts text into a list of token IDs. \\
    & \texttt{\textbf{detokenize}(q, token\_ids) -> str} & Converts token IDs back into text. \\
     & \texttt{\textbf{get\_vocabs}(q) -> list[list[byte]]} & Retrieves the vocabulary list. \\
    
    \midrule
    \multirow{1}{*}{\emph{OutputText} (\emph{Allocate})}
    & \texttt{\textbf{get\_next\_dist}(q, emb) -> future[Dist]} & Gets the next token distribution. \\
    
    \bottomrule
    \end{tabular}
    \label{tab:api}
\end{table*}

The \sys programming model provides an API for developers to create what we call \emph{\programs}—specialized programs that orchestrate LLM generation. 
Each \program executes within a single-threaded, event-driven runtime. Concurrency within a \program is handled through asynchronous, non-blocking API calls, a model well-suited for I/O-bound agentic workflows that spend significant time waiting for API calls or tool execution to complete.

The programming model for \programs addresses three key goals: (1) \emph{expressiveness}—supporting a wide range of generation logic, particularly addressing requirements R1-R3 outlined in \S\ref{sec:introduction}; (2) \emph{efficiency}—enabling performance optimizations through fine-grained resource control; and (3) \emph{extensibility}—remaining agnostic to specific LLM architectures, while providing a foundation for future extensions.

\autoref{tab:api} provides an overview of \sys APIs. Among the 42 APIs, 
18 are core to defining the LLM forward pass and resource management in the inference layer, which we elaborate on in \S\ref{sec:abstractions}. The remaining 24 APIs are for runtime management, inter-\program communication, and I/O, which are essential for building interactive, agentic applications. These APIs do not require GPU processing and are handled entirely by the control layer. We describe them in \S\ref{sec:runtimeapi}.

\paragraphb{Scope and tradeoffs} 
The \sys API is designed for Transformer-based LLMs. It abstracts the workflow into three stages, exposing each as an opaque function accessible through the API. As a result, low-level GPU optimizations, such as kernel fusion and scheduling~\cite{dao2022flashattention}, quantization techniques~\cite{lin2024awq}, or tensor-parallel execution strategies~\cite{zheng2022alpa}, remain orthogonal concerns delegated to the inference layer (\S\ref{sec:arch}).
The \sys API prioritizes fine-grained programmability over simplicity. This incurs programming complexity. We ease the burden through a high-level support library that provides common subroutines and abstractions (\S\ref{sec:support}).

\subsection{Abstractions}
\label{sec:abstractions}

\sys views the LLM forward pass as a three-stage process: (1) \textbf{embed}, which prepares input embeddings from raw data (text, images); (2) \textbf{forward}, which computes output embeddings from input embeddings, potentially utilizing the KV cache; and (3) \textbf{sample}, which derives meaningful outputs (\eg, next-token logits) from output embeddings. APIs related to the model forward pass fall into one of these three categories. \Programs combine API calls from each of these stages to define a complete forward pass.

\paragraphb{Resources}
\sys defines two primary resources that can be passed from stage to stage by pointer.
(1) \embed represents a sequence of token embeddings (input or output), allocated on a per-token basis to offer flexibility in manipulating individual token representations.
(2) \kv represents a contiguous chunk of the KV cache, following PagedAttention~\cite{kwon2023efficient}. A \kv can consist of 8--32 tokens.
\sys requires explicit resource allocation and deallocation (\eg, \code{alloc\_embed} and \code{alloc\_kvpage} in \autoref{tab:api}) by \programs. The returned resource handles are opaque pointers to the underlying memory managed by \sys. Each inferlet has its own virtual resource address space, managed by \sys's control layer. It is possible to share resources between \programs through import/export APIs (\autoref{tab:api}), where the control layer handles the necessary physical-virtual resource address mappings.

\paragraphb{Command queue}
\label{sec:cmdqueue}
Command queue abstracts a logical sequence of API calls issued by an \program. Its purpose is to inform the batch scheduler (\S\ref{sec:batch_scheduler}) to make optimal batching decisions by (1) making the dependencies between API calls unambiguous, and (2) enabling the setting of priorities between different queues.
All API calls that are batch processed by the batch scheduler, such as resource allocation/deallocation, forward pass, and sampling, take a pointer to a command queue object (\cmdqueue) as an argument. On the other hand, API calls that do not require batch processing, such as I/O operations, do not need a command queue argument and are handled directly by the control layer.
For example, in the following \program program, \sys's control layer may batch the two \code{forward} API calls depending on different \cmdqueues so that GPU can process them in parallel in the inference layer.
\begin{examplecode}
q1,q2 = create_queue(), create_queue()
forward(q1, [], iemb1, [], oemb1)
forward(q2, [], iemb2, [], oemb2)
await synchronize(q1)
await synchronize(q2)
\end{examplecode}

\subsection{APIs}

\paragraphb{Embed API}
The embed APIs are responsible for converting raw input data (\eg, text, images) into \embed pointers, which can then be utilized by other APIs. Key functions in this category include \code{embed\_txt} and \code{embed\_img}, which populate allocated \embed pointers based on the provided input data. Additionally, \sys offers auxiliary functions such as \code{tokenize}, \code{detokenize}, and \code{num\_embeds\_needed} to facilitate preprocessing and embedding-related operations in this stage.

\paragraphb{Forward API}
\code{forward} and \code{forward\_with\_adapter} fall under this category. The \code{forward} API executes the core Transformer forward pass. \code{forward\_with\_adapter} allows users to specify LoRA adapters~\cite{hu2022lora} to support fine-tuned models.
The \code{forward} API takes input \embeds and/or input \kvs, performs the attention and transformation steps, and populates output \embeds and/or output \kvs. \code{forward} operates based on explicit sequence positions associated with the resources.
The API either accepts an explicit attention mask in boolean matrix form or infers it from the provided sequence positions. For example, omitting the \kv of a token that precedes the input \embeds effectively masks that token during attention.
This allows \programs to directly manipulate the attention context, essential for techniques leveraging custom attention masks~\cite{xiao2024efficient,beltagy2020longformer,xiao2024efficient} or modular caching~\cite{gim2024prompt}.
We illustrate its usage in the code below. Lines 1-3 generate one output embedding given the 9 tokens (\ie, prefill).
The same operation can be split into two \code{forward} calls, as shown in lines 6-7, using a \kv to store intermediate states.
We note that the \code{forward} API calls in lines 6-7 may be batched together, even though they use the same command queue, which we cover in more detail under \emph{vertical batching} (\S\ref{sec:batch_scheduler}).
\begin{examplecode}
iemb = alloc_emb(9)     # 9 input tokens
oemb = alloc_emb(1)     # 1 output token
forward([], iemb, [], oemb) 
# in examples, 1 KvPage = 1 token
kv = alloc_kvpage(8)          # 8 tokens
forward([], iemb[:-1], kv)  # gen kvpage
forward(kv, iemb[-1:], oemb)# use kvpage
\end{examplecode}

\vspace{-4pt} 
\paragraphb{Sample API}
\sys provides APIs to extract the final outcomes from the output embeddings produced by the forward APIs. 
For instance, \code{get\_next\_dist} takes an \embeds as input and outputs the next-token probability distribution. The \program then controls sampling or other post-processing using the host language. This flexibility is useful for implementing custom sampling schemes~\cite{kuchnik2023validating,kumar2023certifying} and probability-based LLM introspection such as LLM watermarking~\cite{kirchenbauer2023watermark} that require broad access to the underlying token distribution.
Returning the full distribution to the \program may incur significant memory overhead for large vocabularies (\eg, 128K vocabularies in Llama 3). To mitigate this, \sys truncates the distribution to the top-K tokens. K is configurable by the user, with a default of 256 vocabularies.

\paragraphb{Putting it all together}
The primitives described above provide the building blocks for custom generation logic. Below, we implement an autoregressive loop that generates 10 new tokens given the input prompt ``Hello,'' using greedy sampling.
Since only a single command queue is used in this example, the \cmdqueue argument is omitted for brevity.

\begin{examplecode}
prom = tokenize("Hello, ")
tok_limit = len(inp) + 10

# Resource allocation
prom_emb = alloc_emb(len(prom))
gen_emb = alloc_emb(1)
kv = alloc_kvpage(tok_limit)

# Prefill
pos = list(range(len(prom)))         
embed_txt(prom, pos, prom_emb)        
forward([], prom_emb,                   
     kv[:len(prom)], gen_emb)         

# Decode
for i in range(len(prom), tok_limit):
  dist = await get_next_dist(gen_emb) 
  gen = dist.max_index()              
  print(detokenize(gen))      
  embed_txt(gen, [i], gen_emb)        
  forward(kv[:i], gen_emb,               
       kv[i:i+1], gen_emb)            
                                      
# Resource cleanup    
dealloc_emb(prom_emb)
dealloc_emb(gen_emb)
dealloc_kvpage(kv)
\end{examplecode}

The \program programming model exposes fine-grained control similar to OpenGL in GPU graphics programming. Developers who need low-level flexibility can directly orchestrate resources and generation logic, while most users can rely on higher-level abstractions or reusable libraries built atop this foundation. In practice, many applications will only use these libraries or simply run pre-compiled \programs that suit their needs. For instance, \sys's support library (\S\ref{sec:support}) allows users to implement the above autoregressive loop in just three lines of code:
\begin{examplecode}
ctx = Context(model)
ctx.fill("Hello, ")
ctx.generate_until(max_tokens=10)
\end{examplecode}

\subsection{Supporting agentic workflows}
\label{sec:runtimeapi}

\sys provides essential runtime APIs, such as reading command-line arguments (\code{get\_arg}), querying available models (\code{available\_models}), and enabling communication between user (\ie, the client that initiated the \program) and \programs (\code{send}, \code{receive}).
Additionally, \programs can independently perform I/O operations, such as interacting with external servers using networking APIs (\code{http\_get}, \code{http\_post}) or collaborating with other \programs through message-passing APIs (\code{broadcast}, \code{subscribe}). These capabilities enable the creation of dynamic, interactive workflows that seamlessly integrate LLM inference with external data, tools, computations, and other agents.

\subsection{API extensibility}
\label{sec:traits}

LLMs vary in their capabilities (\eg, text-only, multimodal) and may gain new functionalities over time. To provide a stable yet extensible API, \sys defines each set of related operations as a \code{Trait} (similar to traits in Rust). For example, the \code{InputText} trait includes \code{tokenize} and \code{embed\_txt}, while the \code{Forward} trait includes \code{forward}. Specific model implementations then realize one or more traits, and traits themselves can have dependencies (\ie, supertraits). \autoref{tab:api} summarizes the currently defined traits.
Adding support for new modalities (\eg, audio input) or capabilities simply requires defining a new trait and having models implement it. Existing \programs written against older traits continue to work unchanged. \Programs can query the traits supported by a model at runtime using \code{available\_traits} and adapt their logic accordingly.

On the other hand, traits can be used to extend the API set for advanced optimizations that are not possible with the current API sets. For example, schemes like PyramidKV~\cite{cai2024pyramidkv} and AdaKV~\cite{feng2024ada} rely on internal model statistics like token-level attention scores. One can define a variant of the Forward trait, say, \code{IntrospectiveForward}, that returns such statistics to allow \programs to implement custom caching strategies.

\section{System Architecture}
\label{sec:arch}

\begin{figure}[t]
  \centering
  \includegraphics[width=\columnwidth]{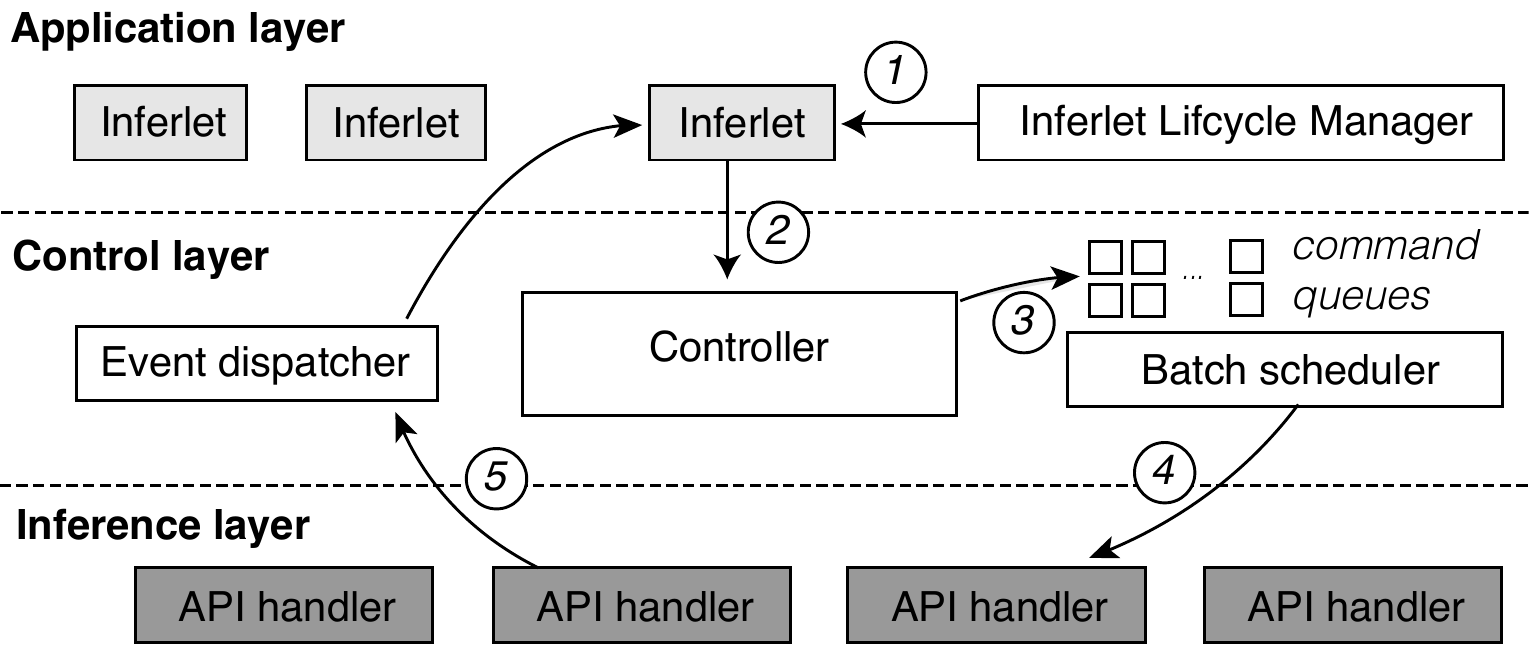}
  \vspace{1pt}
  \caption{\textbf{\Program service workflow (\S\ref{sec:arch})}. The application layer executes \programs that make API calls to the control layer whose batch scheduler adaptively batches these calls and forwards them to the inference layer. Results are sent back to the control layer and then to the \programs.}
  \label{fig:arch}
\end{figure}

\sys employs a three-layer architecture—application, control, and inference—as illustrated in~\autoref{fig:arch}, to efficiently serve multiple user \programs. This design separates concerns: The application layer manages the lifecycle of individual \programs; the control layer oversees system-wide resource coordination, including batch scheduling of API calls; and the inference layer handles low-level, GPU-specific execution of model inference tasks.

\subsection{Application layer}
The application layer executes \programs in a runtime that provides isolation between them. \sys uses WebAssembly (Wasm) runtime to run and sandbox \programs. It serves as an interface for \programs to interact with \sys via the API in \S\ref{sec:api}.
The Inferlet Lifecycle Manager (ILM) in the application layer manages \program lifecycles, including creation, destruction, and communication. It provides RPCs for querying, submitting, and launching \programs with Wasm binaries and arguments (\wcircled{1} in \autoref{fig:arch}). Users can communicate with \programs through the ILM after launch. 
The \code{send} and \code{receive} APIs in \programs handle these user events.

\subsection{Control layer} 
\label{sec:batch_scheduler}
The control layer bridges application and inference layers by batching \sys API calls and routing them to inference layer operations while providing resource abstraction.
A central controller handles API calls from many concurrent \programs (\wcircled{2} in \autoref{fig:arch}), performing three key functions: (1) handling non-GPU API calls (\eg, core runtime API calls and I/Os) directly, (2) managing allocation and virtual address mappings of \embed and \kv resources (\S\ref{sec:abstractions}), and (3) using a batch scheduler to group GPU API calls (\eg, \code{forward}, \code{embed}) for performance efficiency (\wcircled{3}). The event dispatcher receives and broadcasts results from the inference layer back to the \programs (\wcircled{5}).

\begin{figure}[t]
  \centering
  \includegraphics[width=\columnwidth]{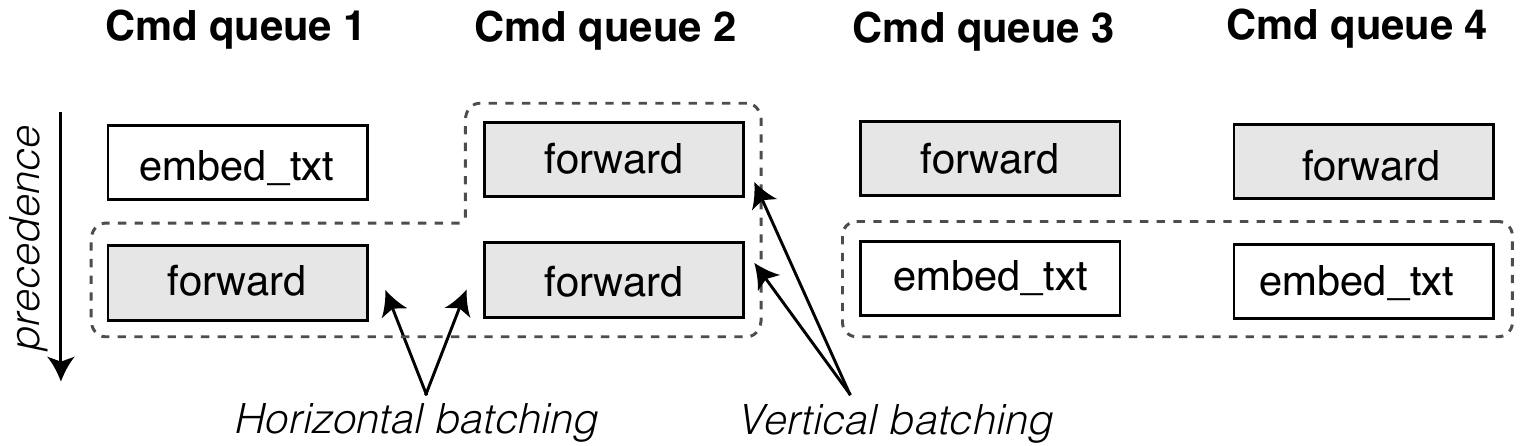}
  \caption{\textbf{Batch scheduling example (\S\ref{sec:batch_scheduler})}. A batch of two \code{embed\_txt} API calls from command queues 3 and 4, or a batch of three \code{forward} API calls from queue 1 and 2, is eligible to be dispatched to the inference layer. Horizontal batching groups calls across different command queues, while vertical batching groups consecutive calls of the same type within the same queue if they do not conflict.}
  \label{fig:batchsched}
\end{figure}

\paragraphb{Resource management}
The control layer manages a global pool of \embed and \kv resources, physically located in the inference layer, with a size configurable at startup based on GPU memory. It provides \programs with a virtualized view of these resources via allocation (\eg, \code{allocate\_kvpage}) APIs to provide resource isolation across \programs. To handle resource contention, the control layer uses a First Come First Serve (FCFS) policy, terminating the most recently created \programs until sufficient resources are freed.

\paragraphb{Batch scheduler}
To enhance GPU utilization and overall throughput, the batch scheduler groups compatible GPU-bound API calls before dispatching them to the inference layer. As shown in \autoref{fig:batchsched}, the scheduler manages pending API calls per command queue (\S\ref{sec:cmdqueue}). By leveraging the unambiguous dependencies and priorities provided by command queues, it forms batches using two techniques:
\begin{enumerate}[leftmargin=*,topsep=0.3em,itemsep=0.02em]
  \item \emph{Vertical batching.} Consecutive commands of the same type within a queue can be batched together, provided they do not conflict, such as writing to the same \kv.
  \item \emph{Horizontal batching.} Commands of the same type across different queues can be batched together.
\end{enumerate}
Within a batch, calls from higher-priority queues are placed earlier. If a batch would exceed the inference backend's maximum supported size, the scheduler truncates from the tail. When multiple API types have eligible batches, the scheduler selects the batch whose oldest pending call has waited the longest.
A central challenge is deciding \emph{when} to dispatch: flushing too early underutilizes kernels; waiting too long inflates latency. \sys uses a work-conserving policy (\S\ref{sec:batch_scheduler_impl}) to maximize GPU utilization .

\subsection{Inference layer}

The inference layer acts as the hardware execution backend for \sys. It receives batched API calls from the control layer via RPC and manages the physical \embed and \kv resources, which are allocated based on the configuration set by the control layer at startup. Importantly, resource management is entirely delegated to the control layer, while the inference layer retains the actual memory.

The inference layer contains multiple API handlers, each specialized for executing a specific type of batched API call (\eg, \code{forward}) that requires GPU execution. Upon receiving a batch (\wcircled{4} in \autoref{fig:arch}), a handler resolves the API call by invoking the necessary GPU-side tensor operations and sends the results back to the control layer's Event Dispatcher (\wcircled{5}). The inference layer resides on the host server connected to the hardware accelerators, and communicate with the control layer using inter-process communication (IPC).

\section{Implementation}
\label{sec:implementation}

\sys is implemented with 13,650 source lines of code (counted with \code{cloc}), including 11,640 lines for the core system and the support library in Rust, and the remaining lines for the GPU-side API handler implementation in Python. \sys uses wasmtime~\cite{wasmtime} as the Wasm runtime to execute \programs and employs WASI to provide system interfaces like network I/O.
The API handlers currently support the Llama family of models.
We highlight two important implementation aspects that enhance system performance, and additionally describe a support library that simplifies development by providing higher-level abstractions for common generation patterns.

\subsection{Adaptive batch scheduling}
\label{sec:batch_scheduler_impl}
We implement the batch scheduler (\S\ref{sec:batch_scheduler}) using a simple work-conserving strategy. The GPU can be in one of two states: \emph{busy} or \emph{idle}. When the GPU is busy, \ie, processing some API calls, the scheduler queues incoming API calls. As soon as the GPU becomes idle, the inference layer immediately notifies the control layer (via IPC) to trigger batch formation.

\subsection{API handlers}
We implement the API handlers of the inference layer using PyTorch~\cite{paszke2017automatic} and the FlashInfer GPU kernel library~\cite{ye2025flashinfer}. These handlers communicate with the control layer via ZeroMQ, a messaging library that supports IPC~\cite{zmq_ipc}.
We also provide a native C++/CUDA implementation of the inference layer that does not depend on PyTorch as part of our open-source release. In our measurements it achieves 10-30\% lower end-to-end latency and more efficient GPU memory utilization than the Python/PyTorch counterpart, owing to custom memory management that pre-allocates and reuses GPU buffers to avoid fragmentation. Because this implementation currently supports only a subset of API traits (e.g., Forward, InputText), we omit a full evaluation here. Interested readers can find further details on the project page.

\subsection{Support library}
\label{sec:support}
We provide a support library, written in Rust, to simplify \program development. 
It provides procedural macros and a lightweight asynchronous runtime to reduce the boilerplate code required to define the Wasm entrypoint and enable the usage of async/await syntax directly in \programs. We offer high-level abstractions, such as \code{Context} for automatically managing \kv, and implement common subroutines such as sampling methods (\eg, top-k, temperature), stopping criteria (\eg, end-of-sequence or maximum tokens), and fork-join parallelism similar to SGLang's API, reducing the need to reimplement these features in each \program. 

\section{Evaluation}
\label{sec:evaluation}

In our evaluation, we address the following questions:
\begin{enumerate}[leftmargin=*,topsep=0.3em,itemsep=0.02em, label=Q\arabic*.]
    \item How does \sys's programming model (\S\ref{sec:api}) facilitate the deployment of emerging LLM applications while meeting requirements R1 to R3 outlined in \S\ref{sec:introduction}? (\S\ref{sec:eval_agentic}-\S\ref{sec:eval_custom})
    \item Can \sys effectively lower the end-to-end latency and improve the throughput of LLM applications compared to existing LLM serving systems? (\S\ref{sec:eval_replicate})
    \item What overhead is introduced by \sys's programmable LLM serving system design? (\S\ref{sec:eval_overhead})
\end{enumerate} 

\paragraphb{Setup} For our evaluation, we utilize a GCP VM G2 instance (\texttt{g2-standard-32}) equipped with an NVIDIA L4 GPU featuring 24 GB of memory to host \sys and baselines.
We use Llama 3 models~\cite{dubey2024llama} (1B, 3B, 8B) with BF16 weights; KV cache and activations also use BF16. We measure end-to-end latency from a remote Python client in a campus network connecting to the serving system on the GCP instance. 

\paragraphb{Baselines} 
We compare \sys with state-of-the-art LLM serving systems: vLLM~\cite{kwon2023efficient} (v0.6.0), SGLang~\cite{zheng2024sglang} (v0.4.4), and specialized frameworks such as LMQL~\cite{beurer2023prompting} (v0.7.3) for structured generation and StreamingLLM~\cite{xiao2024efficient} for attention sink where relevant. To foster a fair comparison focused on architectural differences rather than kernel optimizations, \sys, vLLM, and SGLang all use the FlashInfer GPU backend~\cite{ye2025flashinfer}.

\begin{table}[t!]
    \centering
    \caption{\textbf{LLM applications and inference techniques implemented as \programs}, along with their lines of code (LoC) and compiled Wasm binary sizes. The right column lists support from vLLM (V), SGLang (S), and LMQL (L). The Wasm binary sizes can be reduced by stripping debug symbols. Some applications (\eg, CodeACT) have larger binaries due to embedded libraries such as a JavaScript runtime.
    }
    \vspace{-0.8em}
    \small
    \setlength{\tabcolsep}{2.7pt} 
    \begin{tabular}{@{}lccccl@{}}
        \toprule
        \textbf{Technique} & \textbf{R1-3 (\S\ref{sec:introduction})} &\textbf{LoC} & \textbf{Wasm} & \textbf{Supported} \\
        \midrule
        Support library (\S\ref{sec:support})        & & 728 & -  &        \\
        \midrule
        Text completion       & & 38 & 129 KB   & V, S, L       \\

        \midrule

        ToT~\cite{yao2023tree}                   & R1, R3 & 198 & 148 KB  & S         \\
        RoT~\cite{lee2023recursion}                   & R1, R3 & 106 & 152 KB   &           \\
        GoT~\cite{besta2024graph}                   & R1, R3 & 87 & 171 KB  &           \\
        SKoT~\cite{ning2024skeleton}                  & R1, R3 & 82 & 173 KB  & S        \\
        \midrule

        Prefix caching~\cite{kwon2023efficient}        & R1 & 45 & 131 KB & V, S      \\
        Modular caching~\cite{gim2024prompt}       & R1 & 72 & 139 KB  &           \\
        \midrule
        EBNF decoding~\cite{dong2024xgrammar}         & R2 & 225 & 2 MB   & V, S, L   \\
        \midrule
        Beam search~\cite{freitag2017beam}           & R2 & 98 & 142 KB  & V, L      \\
        Watermarking~\cite{kirchenbauer2023watermark}          & R2 & 43 & 130 KB  &           \\
        Output validation~\cite{kuchnik2023validating}      & R2 & 52 & 131 KB  &           \\
        \midrule
        Speculative decoding~\cite{saxena2023prompt}      & R2 & 255 & 152 KB  & V         \\
        Jacobi decoding~\cite{santilli2023accelerating}       & R2 & 88 & 96 KB   &          \\
        \midrule
        Attention sink~\cite{xiao2024efficient}        & R1 & 60 & 133 KB   &           \\
        Windowed attn~\cite{beltagy2020longformer}.    & R1 & 60 & 133 KB  &         \\
        Hierarchical attn.~\cite{tang2022ast}       & R1 & 42 & 130 KB   &           \\
        \midrule
        Agent-ReACT~\cite{yao2023react}       & All & 60 & 309 KB  &           \\
        Agent-CodeACT~\cite{wang2024executable}       & All & 62 & 6.7 MB &           \\
        Agent-SWARM~\cite{zhuge2024gptswarm}       & All & 95 & 135 KB  &           \\
        \bottomrule
    \end{tabular}
    \label{tab:apps}
\end{table} 

\paragraphb{Applications}
To demonstrate the expressiveness and performance of \sys (Q1, Q2), we implement a wide range of LLM applications using the \sys API and Rust, detailed in~\autoref{tab:apps}. These cover standard techniques, advanced reasoning strategies, and agentic workflows. For baseline comparisons, we replicate the same high-level application logic using Python scripts interacting with the respective system's API server.

\subsection{Serving agentic workflows}
\label{sec:eval_agentic}
Agentic workflows, which integrate LLM inference with external tools or multi-step reasoning, highlight the importance of seamless computation and I/O integration. We implement three representative agents--ReACT (web API interactions), CodeACT (code execution), and Swarm (inter-agent communication)--as \programs, showcasing \sys's ability to co-locate I/O. For comparison, we replicate the same workflows in vLLM and SGLang using Python scripts, applying best-effort optimizations and parallelizing client requests where beneficial for throughput.

\begin{figure}[t]
  \centering
  \includegraphics[width=\columnwidth]{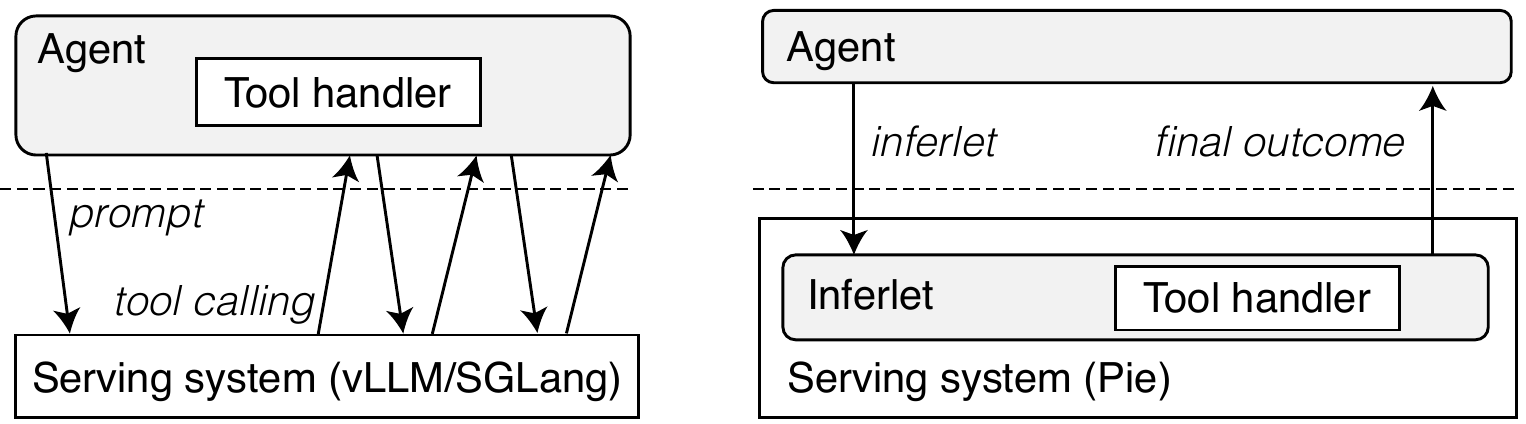}
  \caption{Implementation comparison of agentic workflows in vLLM/SGLang (left) vs. \sys (right).}
  \label{fig:agentic_impl}
\end{figure}

The main differences in implementation between \sys and the baseline systems are illustrated in \autoref{fig:agentic_impl}. In baselines, the application logic only resides within the client, with the serving system handling only LLM inference. This separation incurs latency due to round trips for each external interaction, as well as potential re-prefills when the context changes. In contrast, \sys's \programs encapsulate both inference and external interactions within a single runtime, eliminating these round trips and allowing direct manipulation of the KV cache to retain context across interactions.

\begin{figure}[t]
  \centering
  \includegraphics[width=\columnwidth]{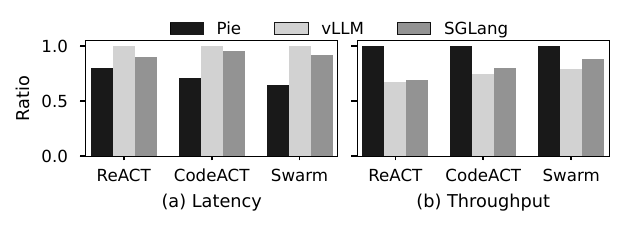}
  \caption{\textbf{Latency} (lower the better) and \textbf{throughput} (higher the better) of LLM agents (X axis: ReACT, CodeACT, and Swarm) hosted by different serving systems (\sys, vLLM, and SGLang). Numbers are normalized to the longest latency or the greatest throughput in each case.}
  \label{fig:agent_latency_tput}
\end{figure}

We measure the latency and throughput in \autoref{fig:agent_latency_tput} using a 1B model and representative workloads involving 8 (ReACT), 8 (CodeACT), and 32 (Swarm) external I/Os, respectively, per agent. The observed latencies are 4.27\,s, 3.18\,s, and 6.14\,s, with corresponding throughputs of 29.94, 40.18, and 5.21 agents/s, respectively.
\sys significantly outperforms baseline systems, \eg, reducing latency by up to 15\% and increasing throughput by up to 30\% on ReACT.
The performance gains are closely tied to the ratio of I/O to the total number of tokens in the workflow.  For instance, we observe no performance difference between baselines when the number of external interactions less than two, and the gap linearly widens as the number of interactions increases.

Two factors contribute to the performance improvements. For smaller models (1B, 3B), the primary factor is the elimination of round-trip latency between the client and server for each external interaction, as the round-trip time (tens of milliseconds) is comparable to token generation time (a few milliseconds). For larger models (8B and above), the dominant factor is the ability to retain the KV cache across external interactions, avoiding costly re-prefills required by the baselines.


\begin{figure}[t]
  \centering
  \includegraphics[width=\columnwidth]{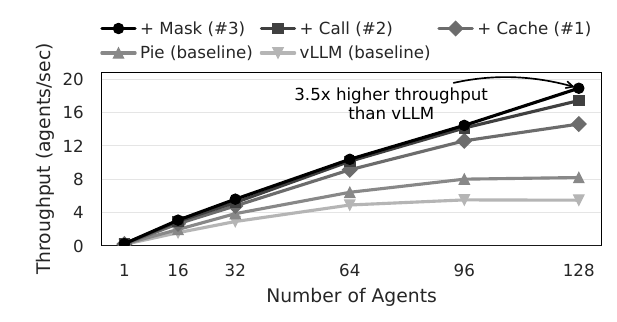}
  \caption{Performance gains via applying workload-specific optimizations to the simple agentic workflow (\S\ref{sec:eval_custom}). Stacked optimizations further improve the performance.}
  \label{fig:stacked_optim}
\end{figure}

\subsection{Customizing generation strategies}
\label{sec:eval_custom}
New LLM generation strategies enhance reasoning, output quality, and efficiency but often require modifications to underlying serving systems to be efficient or even feasible. \sys's programmability addresses this challenge, enabling users to implement and deploy novel strategies as \programs without modifying the core system.

\begin{figure*}[t]
  \centering
  \includegraphics[width=\textwidth]{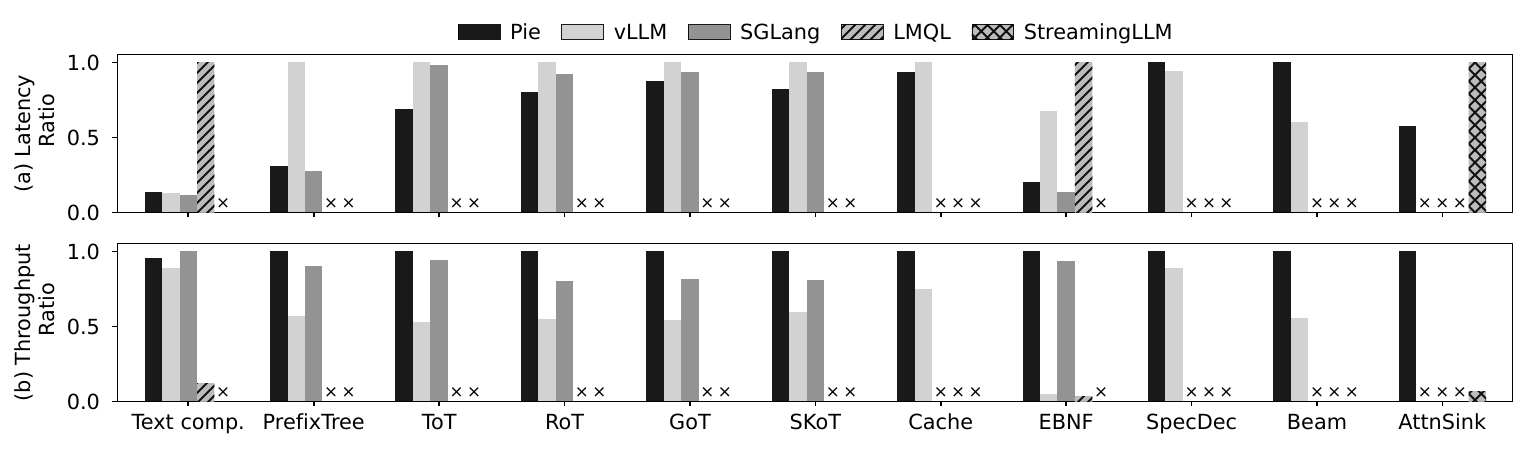}
  \vspace{-4ex}
  \caption{\textbf{Latency} (lower the better) and \textbf{throughput} (higher the better) of example LLM inference techniques hosted by different serving systems.
  Numbers are normalized the same way as in~\autoref{fig:agent_latency_tput}.
  \sys matches state-of-the-art serving systems in performance. $\times$ indicates unsupported techniques or impossible comparisons. For example, SGLang lacks support for the specific speculative decoding implementation~\cite{saxena2023prompt} used by \sys and vLLM.}
  \label{fig:latency_tput_graph}
\end{figure*}
\paragraphb{Applying novel optimizations}
\sys creates new opportunities for application-level inference optimizations.
Consider a typical agentic workflow that initiates multiple LLM function calls to external APIs based on the user query and the provided set of available APIs.
Let us assume that this application uniquely exhibits the following characteristics:
\begin{enumerate}[leftmargin=*]
    \item Certain APIs are used more frequently than others.
    \item The majority of APIs are designed to be fire-and-forget.
    \item Most APIs within a set are invoked only once.
\end{enumerate}
This application-specific knowledge cannot be explicitly leveraged by existing serving systems, which typically rely on implicit optimizations like automatic prefix caching in vLLM.
However, a \sys user can optimize this workflow by employing the following techniques, each exploiting an aforementioned characteristic:
\begin{enumerate}[leftmargin=*, label=\#\arabic*.]
    \item Retain the KV cache of frequently used API documentation using \texttt{export\_kvpage}.
    \item Concurrently call APIs immediately upon detecting the callable signature in the generated tokens.
    \item Drop the KV cache of API specifications used only once from the context using \texttt{mask\_kvpage}.
\end{enumerate}
As \autoref{fig:stacked_optim} shows, each successive optimization builds upon the last, culminating in a 3.5$\times$ throughput increase over the baseline Python-implemented workflow on vLLM. This highlights how \sys's fine-grained control over KV cache (R1), generation process (R2), and I/O (R3) enables developers to tailor inference strategies to application semantics, yielding substantial performance benefits beyond what generic system features like prefix caching can offer alone.

\paragraphb{Deliberate prompting strategies}
We implement four deliberate prompting strategies as \programs: Tree-of-Thought (ToT), Recursion-of-Thought (RoT), Graph-of-Thought (GoT), and Skeleton-of-Thought (SkoT), each comprising approximately 80-100 lines of code (see \autoref{tab:apps}).
In our implementation, we use a simplified version of the tasks described in the original papers (\eg, arithmetic tasks for ToT and RoT, and document summarization for GoT). For RoT, we allow recursive branching up to a depth of 5, resulting in a maximum of $2^5$=32 branches.
\sys achieves better performance across all four strategies, with latency reductions of up to 28\% and throughput improvements of up to 34\% compared to baseline systems.
The performance advantage arises from \sys's program-controlled KV cache reuse, which offers more precise control and flexibility than the implicit KV cache management in existing systems.
This proves beneficial for complex strategies like Recursion-of-Thought (RoT), where SGLang's RadixAttention cannot be effectively applied due to the dynamic nature of the graph structure.
Furthermore, \sys's integrated I/O capabilities provide additional performance benefits for strategies like Tree-of-Thought (ToT), where the system can efficiently interleave compute during value evaluation phases, such as the symbolic evaluation to trim the search space.

\paragraphb{Attention-level techniques}
Leveraging the \code{mask\_kvpages} and mask argument in the \code{forward} API, \sys can implement various attention-level techniques, including attention sink, windowed attention, and hierarchical attention.
To the best of our knowledge, these techniques have not been previously implemented in vLLM or SGLang. We thus compare \sys against specialized implementations, \eg, StreamingLLM for attention sink~\cite{xiao2024efficient}.
Compared to the original StreamingLLM implementation, \sys demonstrates significant performance improvements, achieving 1.5$\times$ lower latency and over 30$\times$ higher throughput. While this comparison is influenced by differences in the underlying GPU kernel libraries, it highlights \sys's ability to effectively express and implement attention-level techniques as \programs.

\subsection{Replicating existing serving features}
\label{sec:eval_replicate}
A key aspect of \sys is its ability to programmatically replicate optimizations often implemented as monolithic, system-wide features in other LLM serving systems. This allows such optimizations to be composable and customized per application. We implement several such features as \programs.

For prefix caching, we replicate vLLM's mechanism using \texttt{export\_kvpage} and \texttt{import\_kvpage} to enable application-controlled cache sharing. For speculative decoding, we implement vLLM's n-gram prompt-lookup method~\cite{saxena2023prompt}. We also implement beam search. Additionally, we create prefix trees (RadixAttention equivalent) to support SGLang-style branching generation by managing shared KV prefixes and divergent streams. Finally, we implement structured generation by integrating a Rust-based constrained decoding library (\texttt{llguidance}) via Wasm to constrain token sampling at each step.
We highlight the EBNF implementation as an example of \sys's ability to integrate third-party libraries seamlessly.

As shown in \autoref{fig:latency_tput_graph}, \sys generally achieves comparable performance to vLLM and SGLang on these tasks. For beam search using three beams, \sys exhibits slightly higher latency, but achieves better throughput. For EBNF decoding using JSON grammar, \sys matches SGLang's performance and significantly outperforms vLLM and LMQL. These results demonstrate that \sys's programming model is both expressive and efficient, enabling the implementation of state-of-the-art optimizations at the application level.

\subsection{System overhead analysis}\label{sec:eval_overhead} 

We employ several microbenchmarks to analyze the performance impact and overhead introduced by \sys's programming model and system architecture.

\begin{figure}[t]
  \centering
  \includegraphics[width=\columnwidth]{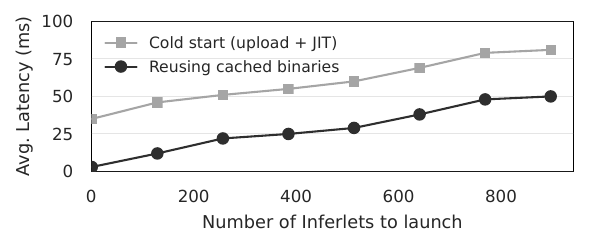}
  \vspace{-4ex}
  \caption{Average latency to launch an \program. The latency is insignificant compared to per-token generation latency. Caching JIT-ed binaries helps reduce latency.}
  \label{fig:launch_overhead}
\end{figure}

\paragraphb{Time to launch new \programs} 
\sys is designed to serve multiple concurrent \programs, each running in its own Wasm runtime instance. The number of concurrent \programs can grow to several hundred. 
This might raise concerns about the overhead associated with launching a large number of \programs. 
To investigate this, we measure the end-to-end time required to launch an \program for the text completion task (in~\autoref{tab:apps}). Specifically, we modify the \texttt{text\_completion} \program to send an acknowledgement message to the user before initiating token generation. We then measure the time elapsed between the request to launch an \program and the reception of acknowledgement, from the Python client.
We measure the latency under two scenarios: cold start and warm start. In a cold start, the client uploads the Wasm binary before launching the \program, and \sys JIT-compiles the received binary. In a warm start, the client utilizes the cached binary in \sys.
As shown in~\autoref{fig:launch_overhead}, the launch overhead ranges from 10 ms to 50 ms for a warm start and 35 ms to 81 ms for a cold start when up to 896 \programs request launching at the same time.
This overhead is insignificant given that the typical per-token generation latency ranges from 10 ms to 60 ms.
\sys achieves the low startup latency through the use of wasmtime's pooled allocation feature~\cite{wasmtime_poolingallocationconfig}, which preallocates virtual memory for the Wasm runtime for up to 1,000 concurrent instances.

\begin{figure}[t]
  \centering
  \includegraphics[width=\columnwidth]{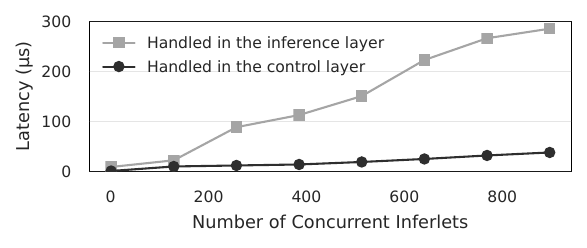}
  \vspace{-4ex}
  \caption{Latency of API calls when handled by the control or inference layer. Higher latency in the inference layer is attributed to the IPC boundary crossing and Python-side API call deserialization overhead.}
  \label{fig:api_overhead}
\end{figure}

\begin{figure}[t]
  \centering
  \includegraphics[width=\columnwidth]{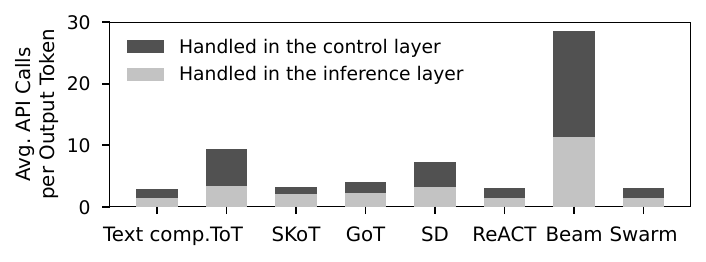}
  \vspace{-4ex}
  \caption{Average number of API calls per task normalized by the number of generated tokens. Beam search (beam=5) involve significantly more API calls because only the tokens from the winning beam are counted as output.
  }
  \label{fig:num_api_calls}
\end{figure}

\paragraphb{Overhead per API call in \programs}
We examine the overhead of API calls since \programs can trigger hundreds to thousands of calls per task, potentially creating performance bottlenecks. API calls fall into two categories: (1) handled by the control layer (\eg, \texttt{get\_arg}, \texttt{send}) and (2) handled by the inference layer (\eg, \texttt{forward}, \texttt{embed\_txt}). \autoref{fig:num_api_calls} shows the frequency of API calls for various tasks, normalized by the number of output tokens generated. For basic text completion, each output token corresponds to approximately 1.6 calls handled by the inference layer and 1.5 by the control layer. A token generated by more complex operations like beam search (width=3) corresponds to significantly more---about 17 and 13 calls handled by the inference and control layer, respectively.
We measure the overhead of each API call in microseconds, as shown in~\autoref{fig:api_overhead}. This overhead is the time from issuing an API call to its completion, excluding handling time (\eg, GPU execution or controller processing, which is typically 6ms-50ms per token on 8B models). We disable batch scheduling for this measurement. The measured latency represents an upper bound since most API calls are executed asynchronously, allowing some latency to be hidden as they do not wait for the previous call to complete.
API calls handled in the control layer are inexpensive, costing less than 30 microseconds per call, even with 896 concurrent \programs. In contrast, those handled in the inference layer are costlier, ranging from 10 to 300 microseconds per call, depending on the number of concurrent \programs. 
The main overhead arises from Python's single-threaded deserialization of API calls, which becomes noticeable with more concurrent \programs. IPC boundary crossing adds minor, constant latency. A native multi-threaded implementation of inference layer (\eg, in C++/Rust) could mitigate this.

\begin{table}[t!]
    \centering
    \caption{\textbf{Opportunity cost of \sys's programming model.}
     Compared to vLLM, the overhead is small compared to \sys's per-token latency (17 ms to 65 ms).
    }
    \vspace{-0.8em}
    \begin{tabular}{lc}
        \toprule
        \textbf{Component} & \textbf{Latency} \\
        \midrule
        Text completion TPOT (vLLM)         & 64.06 ms \\
        \midrule 
        Lack of pipelined sampling on GPU         & +1.320 ms \\
        Lack of pipelined input embedding on GPU  & +0.070 ms \\
        Overhead of control layer batch scheduling         & +0.050 ms \\
        Overhead of returning output distribution & +0.070 ms \\
        Boundary crossing (control-inference layer) & +0.006 ms \\
        Boundary crossing (application-control layer) & +0.001 ms \\
        Wasm processing overhead & +0.001 ms \\
        \midrule
        Text completion TPOT (\sys)         & 65.59 ms \\
        \bottomrule
    \end{tabular}
    \label{tab:opportunity}
\end{table}

\paragraphb{Opportunity cost of the programming model}
The fine-grained APIs in \sys create an opportunity cost when compared to monolithic systems like vLLM. A monolithic decoding loop, such as vLLM's, offers three primary advantages: (1) pipelining the embedding of input tokens and the sampling of output tokens with the LLM forward pass, (2) eliminating the need for separate management of output token distributions, and (3) simplifying batch scheduling.
To measure this cost, we conducted an ablation study on \sys's programming model. We created less granular, custom APIs (\eg, \texttt{forward\_with\_sampling}) that emulate the fused operations of a monolithic system and measured the resulting performance gains. The results from 32 concurrent \programs, summarized in~\autoref{tab:opportunity}, indicate that the opportunity costs associated with advantages (2) and (3) are relatively minor. In contrast, the lack of pipelining—advantage (1)—imposes a more significant latency overhead, ranging from 0.23 to 1.39 ms per token.
Despite these costs, the overhead remains small compared to \sys's typical per-token latency of 5 to 60 ms. This highlights a trade-off: for highly latency-sensitive applications, introducing specialized APIs could boost performance, albeit at the cost of reduced composability.

\paragraphb{Impact of model size}
The relative overhead of \sys diminishes as model size increases, as the system's overhead is amortized over the longer computation time. As shown in \autoref{tab:model}, which compares the end-to-end time per output token (TPOT) with vLLM, the performance gap between the two systems narrows for larger models. For instance, with an 8B parameter model, the 1.53 ms overhead from \sys constitutes just 2.4\% of vLLM's TPOT.

\paragraphb{Batching strategies}
To demonstrate the advantage of adaptive batch scheduling (\S\ref{sec:batch_scheduler_impl}), we compared its throughput against three baseline strategies: no batching (Eager), fixed-size batching based on queue length (K-only), and timeout-based batching based on wait time (T-only). Under a fully saturated scheduler with 128 concurrent \programs, the adaptive strategy proved superior. As shown in \autoref{tab:batch_size}, it delivered up to 17$\times$ higher throughput than the Eager baseline and outperformed the K-only and T-only strategies by 8--40\%.
\begin{table}[t!]
    \centering
    \caption{\textbf{Generation time per output token (TPOT)} for text completion tasks across different model sizes. Larger model sizes help amortize the overhead of \sys.}
    \vspace{-0.8em}
    \begin{tabular}{lcccc} 
        \toprule
        \textbf{Param. size} & \textbf{vLLM} & \textbf{\sys} &\textbf{Overhead (\%)} \\
        \midrule
        8B  & 64.06 ms & 65.59 ms & 1.53 ms (2.39\%) \\
        3B  & 30.30 ms & 32.01 ms & 1.71 ms (5.64\%) \\
        1B  & 16.83 ms & 18.75 ms & 1.92 ms (11.41\%) \\
        \bottomrule
    \end{tabular}
    \label{tab:model}
\end{table}
\vspace{-5pt}
\begin{table}[t!]
    \centering
    \caption{\textbf{Throughput across batching strategies.} The adaptive policy, using work-conserving scheduling, achieves the highest throughput compared to no batching (Eager) and the fixed-rate policies (K-only, T-only).}  \vspace{-0.8em}
    \begin{tabular}{cccccc}
    \toprule
    & \textbf{Eager} & \textbf{K-only}  & \textbf{T-only} & \textbf{Adaptive} \\
    \midrule
    Requests/s & 5.61  & 30.09  & 78.11  & 84.85   \\
    \bottomrule
    \end{tabular}
    \label{tab:batch_size}
\end{table}

\section{Discussion and Limitations}

\paragraphb{Security implications}
Delegating control to user \programs and enabling network I/O expand the attack surface relative to closed-loop serving. Risks include side-channel leakage, resource-exhaustion (DoS), and misuse of access to token distributions (\eg, aiding jailbreaks or model extraction). A comprehensive security treatment is beyond this paper's scope, but hardening is essential for commercial adoption of \sys. Promising directions include capability-based I/O, per-\program rate limits and quotas, and limiting or obfuscating exposure of logits (\eg, top-$K$ caps).

\paragraphb{Runtime implementation choice}
We chose WebAssembly (Wasm) as the \program runtime to balance isolation, startup latency, and portability for multi-tenant workloads. Alternatives fared worse for our goals: native OS processes~\cite{gim2025serve} provide weaker containment; containers (\eg, Docker) add cold-start and resource overhead for frequent \program instantiation; embedded interpreters (\eg, Lua) often incur FFI overhead and provide weaker isolation. In practice, Wasm offered the best trade-off for \sys.

\paragraphb{Scalability of the control layer}
Today, \sys’s control layer is a centralized batch scheduler feeding a single inference backend. Real deployments must span multiple GPU nodes, raising challenges in load balancing, \kv locality (placement/migration), and end-to-end SLO enforcement. Scaling \sys thus requires distributed coordination, globally-aware scheduling, and cache-aware placement policies, as well as fault tolerance across nodes.

\paragraphb{Retrofitting programmability into existing systems}
Supporting fine-grained programmability entails three architectural changes: (1) integrating a runtime to execute user code safely, (2) designing low-level APIs for control and resource management, and (3) redesigning batch scheduling around API-level GPU operations. Retrofitting these into current serving stacks would amount to \sys's architecture. That said, more incremental, coarser-grained programmability layers that better fit existing designs remain an interesting avenue for future work.

\paragraphb{Handling resource contention}
When demand for \kv exceeds capacity, \sys currently applies an FCFS policy that terminates the most recently started \program to free resources. Richer policies could improve fairness and efficiency: per-\program quotas, SLO/priority-aware admission and preemption, and even controlled overprovisioning by swapping \kv between CPU and GPU memory. Exploring these mechanisms is promising future work.

\section{Related Work}
\label{sec:related_work}

\sys builds on a lineage of programmable systems~\cite{engler1995exokernel,bershad1995extensibility,morris1999click,hindman2011mesos,mckeown2008openflow,ebpf2024,peter2014arrakis} across operating systems, networking, and distributed frameworks, which expose fine-grained interfaces to separate application-level control from system internals—a core principle \sys applies to the domain of LLM serving.

\subsection{Application-aware LLM serving}
\sys is in line with the recent trend towards holistic serving system optimizations for LLM workflows~\cite{santhanam2024alto,luo2025autellix,abhyankar2024infercept,jiang2025rago,agrawal2024taming}. 
Notably, recent systems like SGLang~\cite{zheng2024sglang} and Parrot~\cite{lin2024parrot} have introduced programmatic control constructs. SGLang uses primitives like fork/join/gen to optimize shared generation paths via its RadixAttention, while Parrot employs ``semantic variables'' to improve KV cache reuse based on application context. While offering valuable abstractions for structured generation and prompt management (partially addressing R1 and R2), they are still fundamentally constrained by a monolithic generation process. Consequently, they lack fine-grained, direct control over low-level resources (\eg, individual KV cache pages), the sequence of inference steps (\eg, embedding, attention, sampling), or seamless integration of arbitrary I/O within the generation process ---limiting their ability to meet R3. In contrast, \sys provides lower-level, explicit control via its API, enabling customization beyond what these higher-level abstractions permit.
Orthogonally, several works improve the interaction efficiency between LLMs and external tools to address R3---for example, by parallelizing LLM function calls~\cite{openai_function_calling,kim2023llm} or managing the KV cache while awaiting call completion~\cite{yu2025stateful,gao2025fast}. \sys supports such policies as user-defined programs, eliminating the need for internal changes to the serving system.

\subsection{Programming LLM behavior} 
\sys shares conceptual similarities with recent works on ``LLM programming'', which has emerged along two main directions: (1) constrained decoding using formal grammars, and (2) prompt engineering and orchestration frameworks.
In the first direction, there exists LMQL~\cite{beurer2023prompting}, Guidance~\cite{guidance2025}, Outlines~\cite{outlines2025}, XGrammar~\cite{dong2024xgrammar}, and AICI~\cite{moskal2024aici}. These systems allow developers to control generation semantics—typically via EBNF grammars or declarative constraints—within the confines of the standard decoding loop. Notably, AICI uses WebAssembly to allow more flexibility. In contrast, \sys extends this idea by using a Turing-complete language not only to specify what the model should generate, but also \emph{how} generation is performed. That is, \sys \emph{controls the inference process itself}—including tokenization, attention, KV cache management, sampling, and I/O—step by step. This provides a level of programmability that goes beyond output semantics.
In the second direction, frameworks such as LangChain~\cite{langchain2025}, LangFlow~\cite{langflow2025}, DSPy~\cite{khattab2024dspy}, and AutoGen~\cite{wu2023autogen} simplify LLM agent workflows, and prompt optimization, with a focus on developer usability and output quality. \sys is orthogonal and complementary to these systems. It can serve as a backend by running \programs that either manually embed their logic or are generated from their abstractions. This allows such frameworks to inherit \sys's performance and I/O integration.

\subsection{Efficient Transformer inference}
A growing body of work has improved Transformer inference through quantization~\cite{lin2024awq,wang2023bitnet}, parallelism~\cite{zheng2022alpa,aminabadi2022deepspeed}, GPU kernel optimization~\cite{ye2025flashinfer,dao2022flashattention}, and memory management~\cite{lee2024infinigen,ye2024chunkattention,prabhu2025vattention}.
\sys complements these efforts by enabling a programmable LLM generation process that builds on top of such efficiency-oriented techniques.

\section{Conclusion}
\label{sec:conclusion}
 \sys represents a paradigm shift in LLM serving towards programmability. Moving beyond inflexible monolithic designs, \sys decomposes the token generation process into modular API handlers. This allows user-provided programs (\programs) to orchestrate the entire workflow, enabling fine-grained control over KV cache management, custom decoding procedures, and tight integration of external computation and I/O. We show that this programmability allows \sys to implement a wide array of modern LLM techniques, from custom attention patterns to agentic workflows. Crucially, our evaluation confirms that \sys achieves competitive performance on standard tasks while delivering substantial throughput and latency improvements for complex workloads by facilitating targeted optimizations. Our results demonstrate that programmability is key to the performance and flexibility required by the next generation of LLM applications.

\section*{Acknowledgments}
This work is supported in part by National Science Foundation (NSF) Athena AI Institute
(Award \#2112562) and Yale University. The authors thank the reviewers for their constructive comments.

\newpage
\bibliographystyle{ACM-Reference-Format}
\bibliography{abr-short,references}


\end{document}